\documentclass[times,twocolumn,final,authoryear]{elsarticle}

\usepackage{ycviu}
\usepackage{framed,multirow}

\usepackage{amssymb}
\usepackage{latexsym}

\usepackage{url}
\usepackage{xcolor}
\definecolor{newcolor}{rgb}{.8,.349,.1}

\usepackage{textcomp}
\usepackage{stfloats}
\usepackage{url}
\usepackage{verbatim}
\usepackage{graphicx}
\usepackage{multirow}
\usepackage{cite}
\usepackage{subfig}
\usepackage{hyperref}
\usepackage{pifont}
\usepackage{amsmath,amsfonts}
\usepackage{algorithmic}
\usepackage{algorithm}
\usepackage{array}
\usepackage{xspace}
\makeatletter
\DeclareRobustCommand\onedot{\futurelet\@let@token\@onedot}
\def\@onedot{\ifx\@let@token.\else.\null\fi\xspace}
\def\eg{\emph{e.g}\onedot} 
\def\ie{\emph{i.e}\onedot} 
 
\def\etc{\emph{etc}\onedot}

\makeatother

\newcommand{\squeeze}{\vspace{-1em}}
\newcommand{\squeezesm}{\vspace{-0.5em}}

\journal{Computer Vision and Image Understanding}

\begin{document}

\begin{frontmatter}

\title{SImProv: scalable image provenance framework for robust content attribution}

\author[1]{Alexander \snm{Black}\corref{cor1}} 
\author[1]{Tu \snm{Bui}}
\author[2]{Simon \snm{Jenni}}
\author[2]{Zhifei \snm{Zhang}}
\author[2]{Viswanathan \snm{Swaminanthan}}
\author[1,2]{John \snm{Collomosse}}

\address[1]{CVSSP, University of Surrey, UK}
\address[2]{Adobe Research}


\begin{abstract}
We present SImProv -- a scalable image provenance framework to match a query image back to a trusted database of originals and identify possible manipulations on the query. SImProv consists of three stages: a scalable search stage for retrieving top-k most similar images; a re-ranking and near-duplicated detection stage for identifying the original among the candidates; and finally a manipulation detection and visualization stage for localizing regions within the query that may have been manipulated to differ from the original. SImProv is robust to benign image transformations that commonly occur during online redistribution, such as artifacts due to noise and recompression degradation, as well as out-of-place transformations due to image padding, warping, and changes in size and shape. Robustness towards out-of-place transformations is achieved via the end-to-end training of a differentiable warping module within the comparator architecture.  We demonstrate effective retrieval and manipulation detection over a dataset of 100 million images.
\end{abstract}



\end{frontmatter}


\section{Introduction}

Images are a great way to share stories and spread information. However, images can be easily manipulated to tell altered or even completely false stories. As both the number of images shared online each day and the ease of image manipulation grow, the need for tools to provide content provenance information rises. This is addressed in the recently introduced C2PA standards \citep{c2pa} which specifies how provenance information can be encapsulated as meta-data alongside the image content. If an image follows the C2PA standards, users can extract the entire edit story via its secondary stream meta-data. 

This paper addresses a common scenario where meta-data is striped from an image during its online redistribution. It contributes a technique for robustly matching a query (without meta-data) to an original from a trusted database (with full meta-data), followed by an intuitive visualization of the image regions that have been manipulated to differ from the original.

Robust image matching poses many challenges. Images spread online are often subject to {\em benign} transformations such as changes to quality, resolution, aspect ratio, format \etc. Additionally, we aim to match images that have been {\em manipulated} for editorial reasons that alter or falsify their stories (we also call this {\em editorial changes}, as opposed to {\em benign changes}). We note that cryptographic (bit-level) hashes cannot be relied for matching, nor can simple pixel difference operations be used to visualize changes due solely to manipulation. We propose SImProv - a robust and scalable content provenance framework that compliments C2PA. SImProv has two technical contributions:

{\bf Robust Near-Duplicate Image Search.} We learn a visual search embedding that is robust to both benign transformations and content manipulations. We train a convolutional neural network (CNN) using a contrastive learning approach. We use a dataset of original photographs modified in Adobe Photoshop\textsuperscript{TM},  combined with data augmentations simulating benign image modifications. This yields a search embedding for robustly matching a near-duplicate {\em query} image circulating `in the wild' to a trusted database of original images (hereon, we use the term `near-duplicate' to refer to images that undergo certain transformations regardless of such transformations being benign or editorial changes). 

An earlier version of SImProv was proposed at the CVPR workshop on Media Forensics 2021 \citep{ICN}.  The proposed method improves upon this using  instance-level feature pooling methods to improve near-duplicate image search. We show that incorporating these into our image fingerprinting descriptor
improves performance scalability, using a corpus of up to 100 million diverse photographic and artistic images from Behance.Net.  These adaptations demonstrate the utility of our approach for web-scale content authenticity applications.


{\bf Pairwise Image Comparison.} We propose a novel CNN architecture for pairwise image comparison that learns a joint image pair representation. We use this architecture to train two models for near-duplicated detection and editorial change localization respectively. In the near-duplicated detection model, the pair representation is used in conjunction with the individual visual search embeddings of both images to decide whether the two input images are two versions of the same image or completely unrelated {\it distinct} images. In the editorial change visualization model, the pair representation is used to produce a {\it heatmap}  that localizes visual discrepancies  due to editorial manipulation.
The network incorporates both a de-warping and image correlation module, and is trained end-to-end to ignore out-of-place transformation of content e.g. due to padding or warping as well as in-place corruption due to noise.  In this extension of the earlier proposed pair-wise approach \citep{ICN} we show that fusing end-to-end features from the image embedding together with the pair-wise embedding model improves the performance of the near-duplicate detection and re-ranking.  These tasks were previously trained and applied as two sequential, entirely decoupled processes.

\squeeze
\section{Related Work}

The issue of visual content authenticity has been extensively studied from two main perspectives: detection and attribution.

{\bf Detection} typically involves identifying instances of visual tampering or generative content -- `deep fakes' \citep{kaggledf}. This usually requires "blind" detection, where the image in question is the only available information. Different statistical approaches have been explored to localize manipulated regions \citep{zhang2019,ela}. Other methods can identify whether a generative adversarial networks (GANs) was used to create the content \citep{zhang2020, NGUYEN2022103525} or even specify  which GAN, using GAN fingerprints \citep{ganfingerprint}. Most frequently, these methods focus on detection  of fake faces in particular \citep{GUO2021103170, roessler2019faceforensicspp}. 

{\bf Image Attribution} methods aim to link image to data on its provenance, using embedded metadata \citep{cai,origin},  watermarking \citep{hameed2006,devi2009,profrock2006,baba2009}, or perceptual hashing \citep{iscc,dsh2016cvpr,hashnet2017iccv,khelifi2017}. Emerging standards securely transport a cryptographically signed edit history within image metadata \citep{cai,origin,c2pa}.  However, social media platforms often strip metadata from uploaded images, and may even replace it with false information to misattribute the image \citep{strip}. 


CSQ~\citep{csq2020cvpr} approaches hashing as a retrieval/attribution optimization task. Deep Supervised Hashing (DSH)\citep{dsh2016cvpr} and HashNet\citep{hashnet2017iccv} train a siamese  convolutional neural network to learn visual hashes. They use a ranking loss, which is commonly used in visual search \citep{gordo2016deep}. DSDH~\citep{dsdh2017nips} learns metric ranking and classification directly from the hash code.
Our approach uses deep metric learning as well, but differs in that we use contrastive training \citep{chen2020simple} and data augmentation to learn invariances relevant to benign and editorial image transformation. Compared to the image similarity detection challenge and dataset \citep{isc}, which focuses on large-scale retrieval of images subjected to benign transformations, our approach addresses the more complex problem of detecting editorial changes.

{\bf Localization} of image manipulation focuses on identifying image splicing \citep{efros2018} or the use of photo-retouching tools \citep{zhang2019} in blind detection tasks. We tackle the problem by combining perceptual hashing and pair-wise comparison. Our image comparator, the second contribution of this paper, assumes that a trusted "original" image can be found using visual search (the first contribution of this paper). The comparator is able to ignore differences caused by benign image transformations but is sensitive to editorial manipulations. This is achieved through the use of a differential optical flow \citep{raft2020} and dewarping module in our two-stream architecture. Two-stream networks have previously been used to predict the types of editing operations applied to pairs of images  \citep{jenni2018}. Our approach is different because it produces a heatmap of editing operations that is insensitive to specific transformation classes. Another feature of our method is a classification score that can be used at inference to determine whether an image is a benign or manipulated version, or a different image altogether.

\begin{figure*}[ht]
    \centering
    \begin{tabular}{c}
        \renewcommand{\thesubfigure}{}
        \subfloat[]{%
          \includegraphics[width=0.5\linewidth]{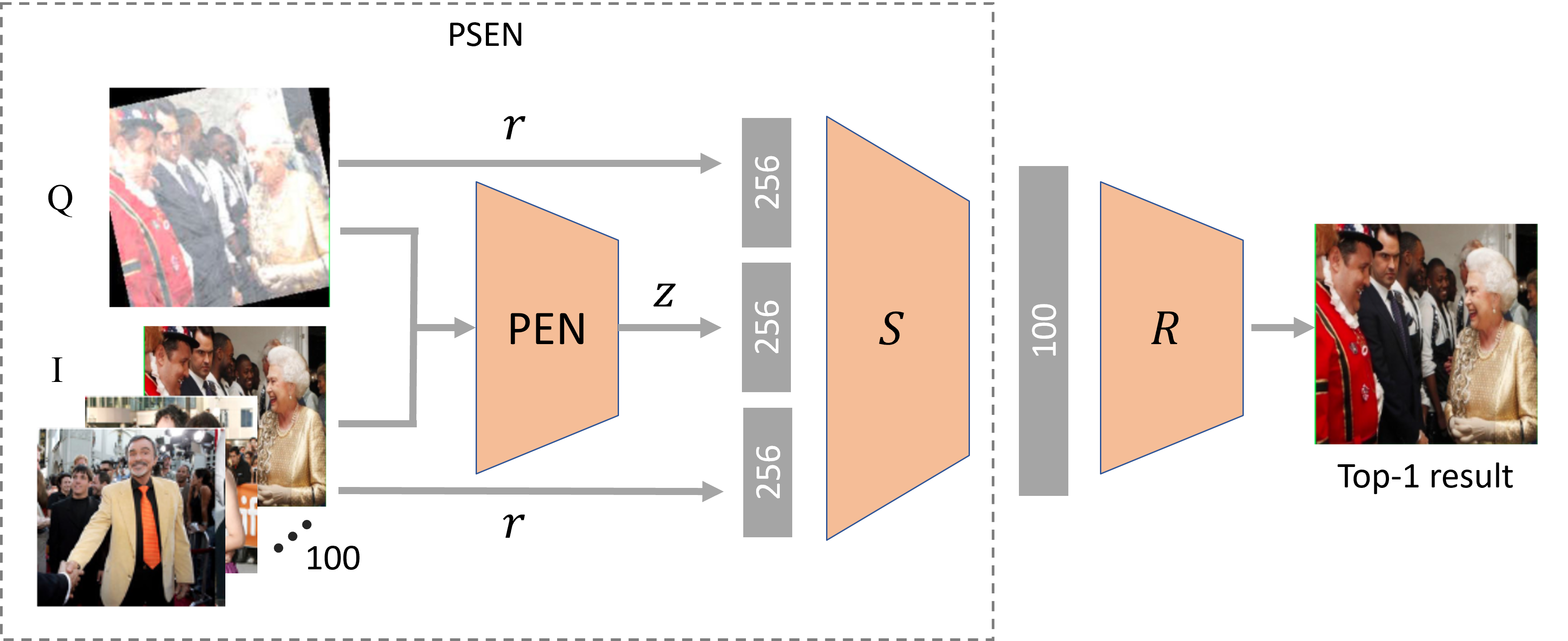}%
          \label{fig:arch_stage_2}
        }\renewcommand{\thesubfigure}{}
        \subfloat[]{%
          \includegraphics[width=0.5\linewidth]{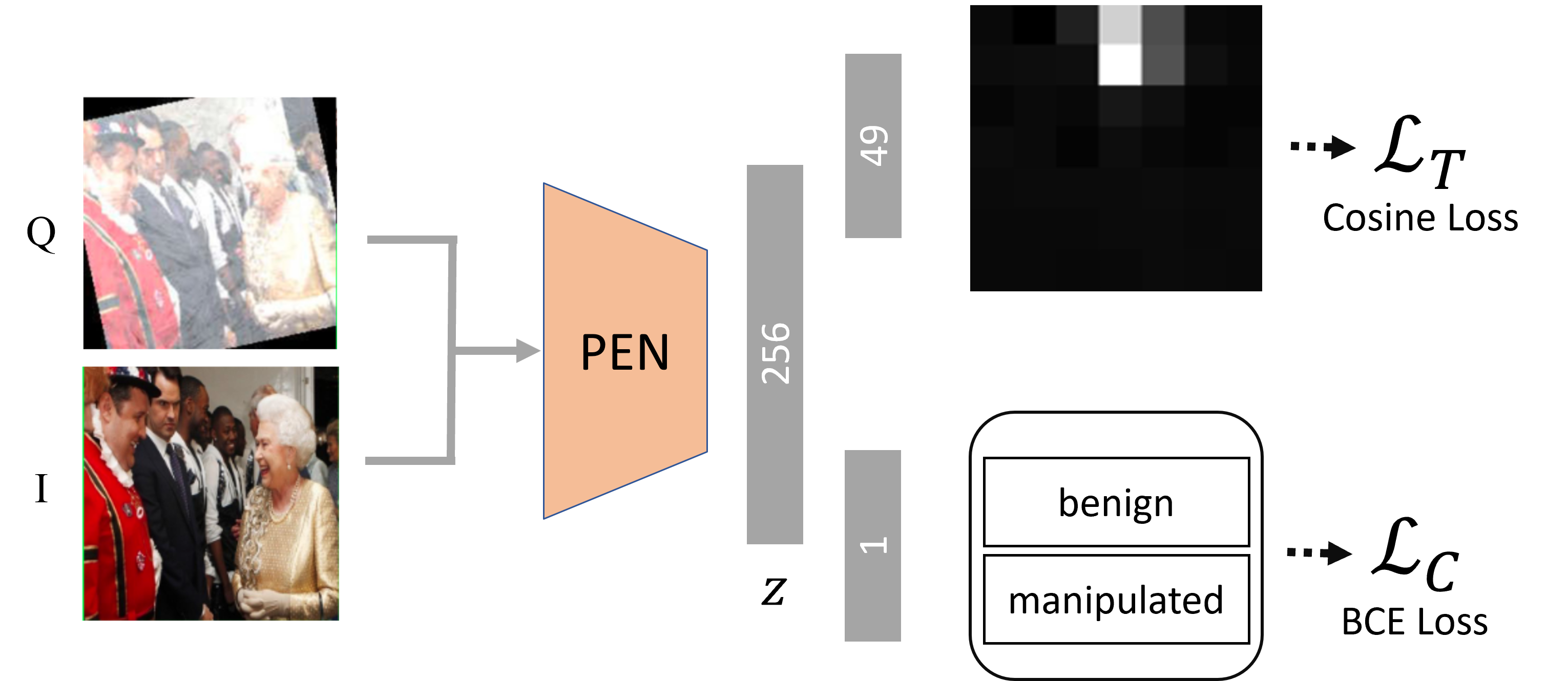}%
          \label{fig:arch_stage_3}
        }
    \end{tabular}
    \caption{Architecture diagrams of each of second and third stages of the proposed SImProv framework. Stage 2 (left) performs re-ranking of the top-100 retrieved results, utilizing a pairwise embedding $z$ to re-order the results and identify a correct match to the query. Stage 3 (right) uses the Pairwise Embedding Network (PEN) to identify whether the query image has been manipulated and localizes the manipulation with a heatmap.}
    \label{fig:arch_main}
    \squeeze
\end{figure*}

\squeeze
\section{Method}

Our approach for image provenance assumes the existence of a trusted database $\mathcal{D}=\{I_1,I_2,...,I_N\}$ containing N original images and their associated provenance information (\eg curated by a trusted publisher, or via a decentralized immutable data-store such as a blockchain). Given a query image $q$, our goals are: (i) determining whether there exists an original version of $q$ in $\mathcal{D}$; and (ii) localizing editorial changes if a match is found. The two goals appear to conflict each other since the former requires robustness to both benign and editorial changes while the latter should be sensitive to editorial manipulations. Learning a single model to achieve our goals is therefore extremely challenging. We instead propose a multi-stage framework. Our SImProv consists of 3 stages: (i) a visual search stage followed by (ii) re-ranking and near-duplicated detection, and finally (iii) detection and visualization of editorial changes (Fig.~\ref{fig:arch_main}). 

Firstly, in \ref{method:search} we describe the representation learning process, used for near-duplicate image search (stage 1).  We develop a model that learns 256-D representations of images that are further binarized into a 128-bit hash for scalable search \citep{faiss2019}. The search is used to identify the most similar images to a users' query image.

Secondly, \ref{method:comparator} describes the Pairwise Embedding Network (PEN) to deliver a pairwise representation of the query and a candidate image. PEN is our core design for the later stages of SImProv.  In \ref{method:reranking}, PEN is integrated to our stage-2 Pairwise Similarity Evaluation Network (PSEN) to re-rank the top k images (k=100) and identify the likelihood of candidate image being a near-duplicated version of the query, as opposed to being just another distinct image.

Finally, \ref{method:loc} describes how PEN is leveraged to identify whether the query image is a manipulated or benignly transformed version of the original (stage 3). If the query is identified as manipulated, we visualize a heatmap of the manipulated region on top of the image.

\squeeze
\subsection{Near-Duplicate Image Search}\label{method:search:representation}
We train a CNN model $f_r(.)$ to encode an image $I$ into a compact embedding space $r=f_r(I) \in \mathbb{R}^{256}$. Additionally, we perform KMeans quantization with 1024 clusters, followed by product quantization, resulting in a 128-bit descriptor of the  256-dimensional embedding \citep{faiss2019}. We employ the ResNet50 \citep{he2016deep} backbone for $f_r$, replacing the final layer with a 256-D fully connected (fc) layer as the embedding. We use DeepAugMix \citep{hendrycks2020} as the pretrained weight. We  finetune with a multiple-positives contrastive loss as described in \citep{ICN}.

During inference, we find our model benefits from geometric pooling (GeM) \citep{gem} in two ways. Firstly, model becomes resolution agnostic and can take larger images as input, capturing more information. Secondly, it allows to focus on local features, which is more beneficial for matching out of place transformed images.

For a set of $K$ spatial feature map activations $\Phi = [\phi_1, \ldots \phi_K]$, the GeM \citep{gem} pooling operation $G$ is defined as $G(\Phi) = [g_1, \ldots g_k, \ldots g_K]$; $g_k = (\frac{1}{|\phi_k|}\sum_{x\in\phi_k}x^{p_k})^{\frac{1}{p_k}}$,  where $p_k$ is a hyper-parameter ($p_k$ is fixed at a default value of $3$ in our experiments).

\subsection{Pairwise Embedding Network}\label{method:comparator}

\begin{figure*}[ht]
\centering
\includegraphics[width=0.9\linewidth]{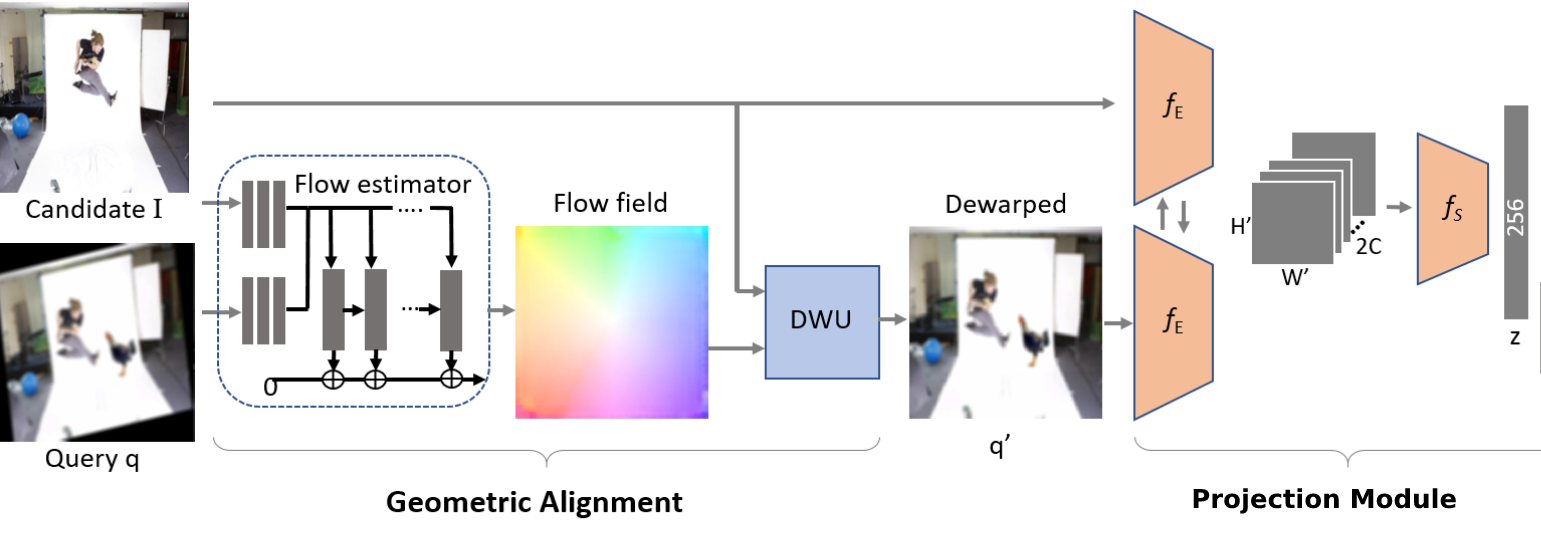}
\caption{Architecture of the proposed Pairwise Embedding Network (PEN).  A candidate match to the user queried image is obtained from near-duplicate search (\ref{method:search}, not shown).  Image alignment is performed via differentiable de-warping unit (DWU) based on a dense optical flow estimate provided by the flow estimator.  The resulting image pair are separately encoded via a feature extractor $f_E(.)$ and the concatenated features passed through $f_S(.)$ to obtain the combined feature $z$.}
\label{fig:comparator_arch}
\squeeze
\end{figure*}

We propose an Pairwise Embedding Network (PEN) that learns a joint representation of two input images (Fig.~\ref{fig:comparator_arch}). This architecture is later utilized for two purposes: near-duplicated detection (\ref{method:sim}) and localization of editorial change (\ref{method:loc}), which corresponds to stage 2 and 3 of SImProv respectively. 


The PEN accepts a pair of query-candidate images as input and outputs a $n$-dimensional ($n=256$ in our experiments) representation of the image pair. The PEN architecture consists of 2 modules: a geometrical alignment module, $\mathcal{F}_A$, followed by a projection module, $\mathcal{F}_P$ (Fig.~\ref{fig:comparator_arch}). Below we describe our designs for $\mathcal{F}_A$ and $\mathcal{F}_P$.

{\bf Geometric Alignment Module} is used to account for the fact that the query $q$ may undergo through geometric transformations which alter the pixel placement. We correct its alignment prior to joint representation learning. In this work we use RAFT \citep{raft2020}, but any flow estimation network that can be trained end-to-end is suitable. 

Our DWU then uses the predicted optical flow to align the query image with the candidate image:
\begin{align} 
    M\; : \; (x,y) &\mapsto (x+\rho^x(x), y+\rho^y(y)) \\
    \textrm{DWU}\,(q|\rho^x, \rho^y) &= \mathcal{S}(M) \in \mathbb{R}^{H\times W}
\end{align}
where $(x,y)$ refers to the pixel coordinates in the query image $q$ which are mapped to their corresponding coordinates $M$ in the candidate image according to the optical flow ${\rho^x, \rho^y}$. $\mathcal{S}(.)$ is a bilinear sampler that effectively fits a local grid around $M$: $\mathcal{S}(M)={M+\Delta M| \Delta M \in \mathbb{R}^2, \left|\Delta M\right|<=1}$ where output coordinates are computed through linear interpolation.

{\bf Projection Module} takes the candidate $I$ and the aligned query, $q'=\mathcal{F}_A(q|I)$ and outputs a single feature $z$. We first extract local features of each image using a shared CNN module: $z_q=f_E(q');\; z_I=f_E(I) \in \mathbb{R}^{H'\times W' \times C}$, where $H'$, $W'$ and $C$ are the new height, width and feature dimension respectively. Our feature extractor $f_E(.)$ is a 3-layer convolutional neural network (CNN) with ReLU activations, batch normalization, and max pooling layers. It outputs features at $\frac{1}{4}$ resolution ($H'=H/4, W'=W/4$ and we set $C=128$). The combined features are then fed into another CNN to learn a fusion representation $z = f_S([z_q, z_I]) \in \mathbb{R}^{256}$,  where $[,]$ is concatenation, and $f_S(.)$ is made up of 4 ResNet residual blocks \citep{he2016deep} followed by average pooling and a fully-connected (FC) layer that outputs 256-dimensional features. The PEN output is used for re-ranking, near-duplicate detection, and manipulation localization, as described below.

\subsection{Near-duplicate detection and Re-ranking}
\label{method:reranking}
Our near-duplicate image search method is designed to produce compact descriptors to enable interactive speeds in search through millions of images. However, the increase in speed comes at a cost of precision. The correct image could end up near the top of retrieval results, but in many cases might not in the first place. 
We propose a re-ranking model based on pairwise comparison of the query image with each of the top-k retrieval candidates. Such pairwise comparison is much slower and is not feasible for search through millions of images, but allows to identify the most likely match within a shortlist. 
The re-ranking  consists of two steps: pairwise similarity evaluation and final reordering. Similarity evaluation produces a similarity confidence score for each of the 100 query-candidate image pairs. Re-ordering looks at the full list of 100 confidence scores and decides which of the candidate images is the most likely match to the query. On  a single GTX 1080 Ti GPU initial search takes 40ms and reordering 400ms on average. Below we describe similarity evaluation (\ref{method:sim}) and reordering (\ref{method:reordering}).

\subsubsection{Image Similarity Evaluation}\label{method:sim}
We propose a Pairwise Similarity Evaluation Network (PSEN) that uses two images as input: the query image $q$ and a candidate image $c$, retrieved by the near-duplicate search model (\ref{method:search}). The PSEN uses the previously obtained individual embeddings of the images, as well as a PEN joint embedding learned from stack of two images together (Fig.~\ref{fig:arch_stage_2}). 

The final output of the model is a confidence score $s$, indicating the likelihood that the query and candidate images are the same image under different transformations:
\squeezesm
\begin{equation}
    s = S([f_r(q), f_r(c), PEN([q, c])]) \in [0, 1]
    \label{eq:S}
\end{equation}
where $[,]$ is concatenation, $f_r(.)$ is the search embedding (\ref{method:search:representation}), $PEN([., .])$ is the pair representation (\ref{method:comparator}) and $S(.)$ is a binary classification fully connected layer. The model is trained with binary cross-entropy loss.

\subsubsection{Re-ordering}\label{method:reordering}
The role of re-orderer is to decide which of the candidate images is the most likely match to the query, based on two pieces of information: the initial raking of near-duplicate image search (distance between candidate and query embeddings) and similarity confidence scores. The index $n$ of the most likely match is defined as $n = R(s_0, \ldots, s_{99}) \in [0, 100]$, where $s_i = S([f_r(q), f_r(c_i), PEN([q, c_i])])$ is the similarity confidence score between the query image and $i$-th retrieved candidate image; $R(.)$ is a neural network, consisting of three fully connected layers of sizes $[8192, 1024, 101]$, respectively.
Re-orderer is trained as a 101-way classifier with cross-entropy loss. First 100 classes correspond to indices of candidate images and the final class indicates that the correct match to the query is not present within the candidate images.

\subsection{Detecting and Localizing Editorial Change}\label{method:loc}

This stage assumes a near-duplicated image to a query has been found after phase 2 (\ref{method:reranking}). In order to predict the benign-manipulated relationship and visualize the possible manipulated regions, we train a second PEN model using a combination of two different loss functions (Fig.~\ref{fig:arch_stage_3}). The first loss is a binary cross-entropy loss $\mathcal{L}_C(.)$ that predicts whether the pair is benign (\ie, the query image $q$ is either identical to or a benign transformed version of the candidate image $I$) or manipulated (\ie, $z$ is a manipulated version of $I$). The second loss function is used to minimize the cosine distance $\mathcal{L}_T(.)$ between the manipulation heatmap derived from $z$ and the ground truth heatmap. This heatmap is generated at a resolution of $t\times t$ using a fully-connected layer $E_t(z) \in \mathbb{R}^{t^2}$.

The total loss is $\mathcal{L}(.) = w_c \mathcal{L}_C(.) + w_t \mathcal{L}_T(.)$ where loss weight $w_c=w_t=0.5$ is set empirically.

\squeezesm
\section{Experiments and Discussion}


\subsection{Datasets}\label{sub:data}
The PSBattles \citep{psBattles} dataset is used to train and evaluate our models. We use the splits and edit localization annotations from \citep{ICN}. The dataset is split into a training set (\textbf{PSBat-Train}) and a test set (\textbf{PSBat-Test}), with 6,364/21,197 and 807/2,960 original/manipulated images in the training and test sets, respectively. The PSBat-Train set is used to train our image retrieval, similarity evaluation, and edit localization models, while the PSBat-Test set is used for the two benchmarks we evaluate.

Additionally, we also evaluate the retrieval efficacy of SImProv on BAM-100M, a large scale dataset consisting of 100M artworks from Behance. BAM-100M is significantly larger and more diverse than ImageNet, since its collection spans many fields beyond photography, such as paintings, graphic designs, advertising and graffiti. We note this is the largest experiment in term of dataset size for image provenance to date. We create two query sets, BAM-Q-Res and BAM-Q-Aug from 1K images sampled at random from BAM-100M. To make BAM-Q-Res, we downscale images at random ratio in range 0.1-0.9 (up to 10x downscaling) with bilinear interpolation keeping aspect ratio. To make BAM-Q-Aug, we apply the same augmentation strategy as in PSBat-Ret.  

\subsection{Metrics}
To evaluate near duplicate search, we measure the ratio of queries that return the relevant images within the top-$k$ retrieval using the Instance Retrieval IR@$k$ metric. Formally, $IR@k = \frac{1}{Q}\sum_{i=1}^Q{\sum_{j=1}^k{r(q_i,j)}}$ where Q is number of queries, relevance function $r(q_i,j)=1$ if the returned image at rank $j$ is relevant to the query $q_i$ (there is only one such image in PSBat-Ret), otherwise 0.

We use Average Precision (AP) to measure the accuracy of both classifiers: the same/different similarity evaluation network and benign/manipulated classifier branch of the edit localization network.

For the generated heatmap, we compute Intersection over Union (IoU) with the ground truth by up-sampling the 7x7 heatmap to the image resolution $H\times W$, converting it to binary with a threshold, and then calculating the intersection and union. This is expressed as $\mathrm{IoU} = \frac{1}{Q}\sum_{i=1}^Q{\frac{S(U_i) \cap T_i}{S(U_i) \cup T_i}}$ where $T_i$ is the $H\times W$ binary ground truth heatmap, $U_i$ is the predicted heatmap after interpolation and thresholding, and $S(U_i)$ is the set of values in $U_i$. We improve the heatmap using the image pair classification result, setting $S(U_i) = U_i$ if the query is classified as manipulated, ${0}^{H\times W}$ if benign, and ${1}^{H\times W}$ if distinct.

\subsection{Evaluating Near-Duplicate Search}
\label{sec:experiment:search}
We compare our retrieval method (both before and after re-ranking) against 9 baselines. {\bf ICN} \citep{ICN} is our initial workshop version of SImProv. \textbf{ImageNet fine.} and \textbf{MSResNet fine.} are the finetuned models on PSBat-Train using our training strategy. All methods produce 128-bit hash code except pHash (64-bit).

\begin{table*}
\caption{Retrieval performance (on 2M images, PSBat-Ret) reported as IR score at ranks [1,10,100], for query images subjected to benign transforms, manipulation, or both. Stage 1 refers to nearest-neighbor search only.}
\centering
\resizebox{\textwidth}{!}{%
\begin{tabular}{l|ccc|ccc|ccc|ccc}
\multirow{1}{*}{\ttfamily Method} & \multicolumn{3}{c|}{\ttfamily Benign} & \multicolumn{3}{c|}{\ttfamily Manip} & \multicolumn{3}{c}{\ttfamily Manip+Benign} & \multicolumn{3}{|c}{\ttfamily Average} \\
& \ttfamily IR@1 &  \ttfamily IR@10 & \ttfamily IR@100  & \ttfamily IR@1 & \ttfamily IR@10 & \ttfamily IR@100  & \ttfamily IR@1 & \ttfamily IR@10 & \ttfamily IR@100 & \ttfamily IR@1 & \ttfamily IR@10 & \ttfamily IR@100   \\ 
\hline \hline
SImProv (stage 1 + 2) & \bf0.9746 & \bf0.9849 & 0.9849 & \bf0.9170 & \bf0.9564 & \bf0.9564 & \bf0.9142 & \bf0.9271 & \bf0.9271 & \bf0.9353 & \bf0.9561 & \bf0.9561  \\
SImProv (stage 1) & 0.9450 & 0.9749 & 0.9849 & 0.8838 & 0.9307 & \bf0.9564 & 0.8064 & 0.8845 & \bf0.9271 & 0.8784 & 0.9300 & \bf0.9561 \\
\hline
ICN \citep{ICN} (stage 1 + 2) & 0.9423 & 0.9811 & \bf0.9867 & 0.9154 & 0.9453 & 0.9453 & 0.8705 & 0.9099 & 0.9164 & 0.9094 & 0.9454 & 0.9494 \\
ICN \citep{ICN} (stage 1) & 0.9305 & 0.9725 & \bf0.9867 & 0.8557 & 0.9061 & 0.9453 & 0.7662 & 0.8582 & 0.9164 & 0.8508 & 0.9123 & 0.9494 \\
MSResNet fine. & 0.7532 & 0.7931 & 0.8258 & 0.9199 & 0.9500 & 0.9655 & 0.6754 & 0.7406 & 0.7884 & 0.7828 & 0.8279 & 0.8599\\
ImageNet fine. & 0.7709 & 0.8988 & 0.9577 & 0.7791 & 0.8689 & 0.9159 & 0.5641 & 0.7348 & 0.8420 & 0.7047 & 0.8342 & 0.9052\\ 
DeepAugMix~\citep{hendrycks2020} & 0.6944 & 0.7743 & 0.8336 & 0.8956 & 0.9385 & 0.9615 & 0.5629 & 0.6725 & 0.7553 & 0.7176 & 0.7951 & 0.8501 \\
CSQ \citep{csq2020cvpr} & 0.1390 & 0.1803 & 0.3292 & 0.2095 & 0.2584 & 0.4122 & 0.0628 & 0.0957 & 0.2214 & 0.1371 & 0.1781 & 0.3209 \\ 
pHash \citep{phash} & 0.3674 & 0.3731 & 0.3768 & 0.3693 & 0.3753 & 0.3764 & 0.1739 & 0.1821 & 0.1857 & 0.3035 & 0.3102 & 0.3130\\ \hline

\end{tabular}}
\squeeze
\label{tab:retrieval}
\end{table*}

Tab.~\ref{tab:retrieval} compares retrieval performance. The two online hashing methods, CSQ~\citep{csq2020cvpr} and HashNet~\citep{hashnet2017iccv}, are among the worst performers. CSQ and HashNet struggle to cope with strong ImageNet-C transformations present during training and test, resulting in lower performance than the classical pHash. ImageNet~\citep{he2016deep}, MSResNet~\citep{msresnet} and DeepAugMix~\citep{hendrycks2020} perform strongly on the Manip set but poorly when they undergo benign transformations. When trained via our contrastive loss (eq.~\ref{eq:simclr}), all models gain with our proposed {\it SImProv (stage 1)} achieving 25\% improvement on Benign IR@1 and 24\% on Manip+Benign versus the pretrained DeepAugMix model. {\it SImProv (stage 1)} also outperforms the finetuned ImageNet/MSResNet by a large margin on all top-k scores and query sets. The improvement of SImProv compared to the one presented in the workshop paper ({\it ICN (stage 1)}) \citep{ICN} can be attributed to geometric pooling, which allows higher resolution input. We demonstrate significant performance improvement of the proposed re-ranking method ({\it SImProv (stage 1+2)}) compared to the naive re-ranking approach used in {\it ICN (stage 1+2)}. 



\subsection{Feature Pooling}
\begin{figure}[t!]
    \centering
    \includegraphics[width=0.85\linewidth]{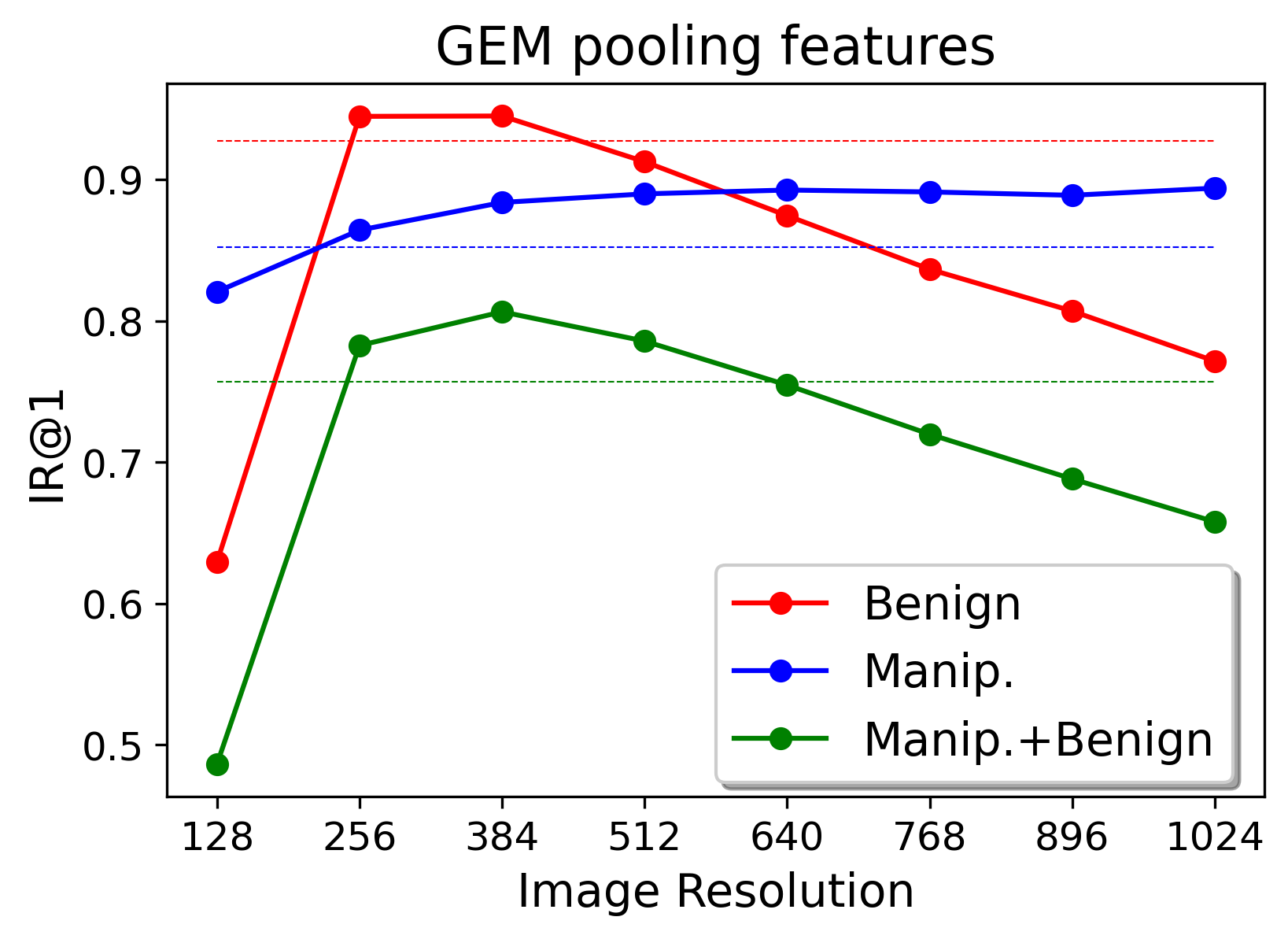}
    \caption{Top-1 performance of GeM features on different input image resolutions. Dash-lines represent the performance without using feature pooling.}
    \squeeze
    \label{fig:resolution}
\end{figure}

We show the superiority of GeM features against the traditional output features from the last FC layer of the retrieval model \citep{ICN} as well as its dependence on input image resolution in Fig.~\ref{fig:resolution}. GeM features work best at 384x384 resolution, outperforming \citep{ICN} by 4\% on the challenging Manip.+Benign test set. The presence of benign transformations hampers GeM performance as the resolution increases, underperforming \citep{ICN} from 512x512 resolution on Benign set and from 640x640 on the Manip.+Benign set.  

Tab.~\ref{tab:pool} compare GeM with \citep{ICN} and a similar feature pooling method - RMAC \citep{rmac}, at two pooling levels, L=3 and L=4. It can be seen that the pooling level does not affect much the performance of both GeM and RMAC. Additionally, RMAC is comparable to GeM, slightly outperforming GeM on the Benign set at L=3 but underperforming on the Manip. set at L=4. However, we choose GeM as the proposed method since it is significantly faster than RMAC. It takes RMAC $28.72$ seconds to perform $1000$ iterations, while GeM is  $\sim 18$ times faster, with just $1.63$ for the same setup.

\begin{table}[t]
\caption{Top-1 retrieval performance of GeM and RMAC features and output from the last FC layer  on the 384x384 resolution test sets.}
\resizebox{\columnwidth}{!}{%
\begin{tabular}{c|ccccc}
 & \multicolumn{2}{c}{RMAC} & \multicolumn{2}{c}{GeM} & CNN \\
 & L=3 & L=4 & L=3 & L=4 &  \\ \hline 
Benign & 0.9503 & 0.9444 & 0.9450 & 0.9441 & 0.9272 \\
Manip. & 0.8777 & 0.8777 & 0.8838 & 0.8926 & 0.8520 \\
\multicolumn{1}{l}{Manip.+Benign} & \multicolumn{1}{l}{0.8025} & \multicolumn{1}{l}{0.8027} & \multicolumn{1}{l}{0.8064} & \multicolumn{1}{l}{0.8068} & \multicolumn{1}{l}{0.7570}
\end{tabular}}
\label{tab:pool}
\end{table}


\subsection{Evaluating Localization of Editorial Changes}

\begin{figure}[t!]
    \centering
    \includegraphics[width=1.0\linewidth,height=5.5cm]{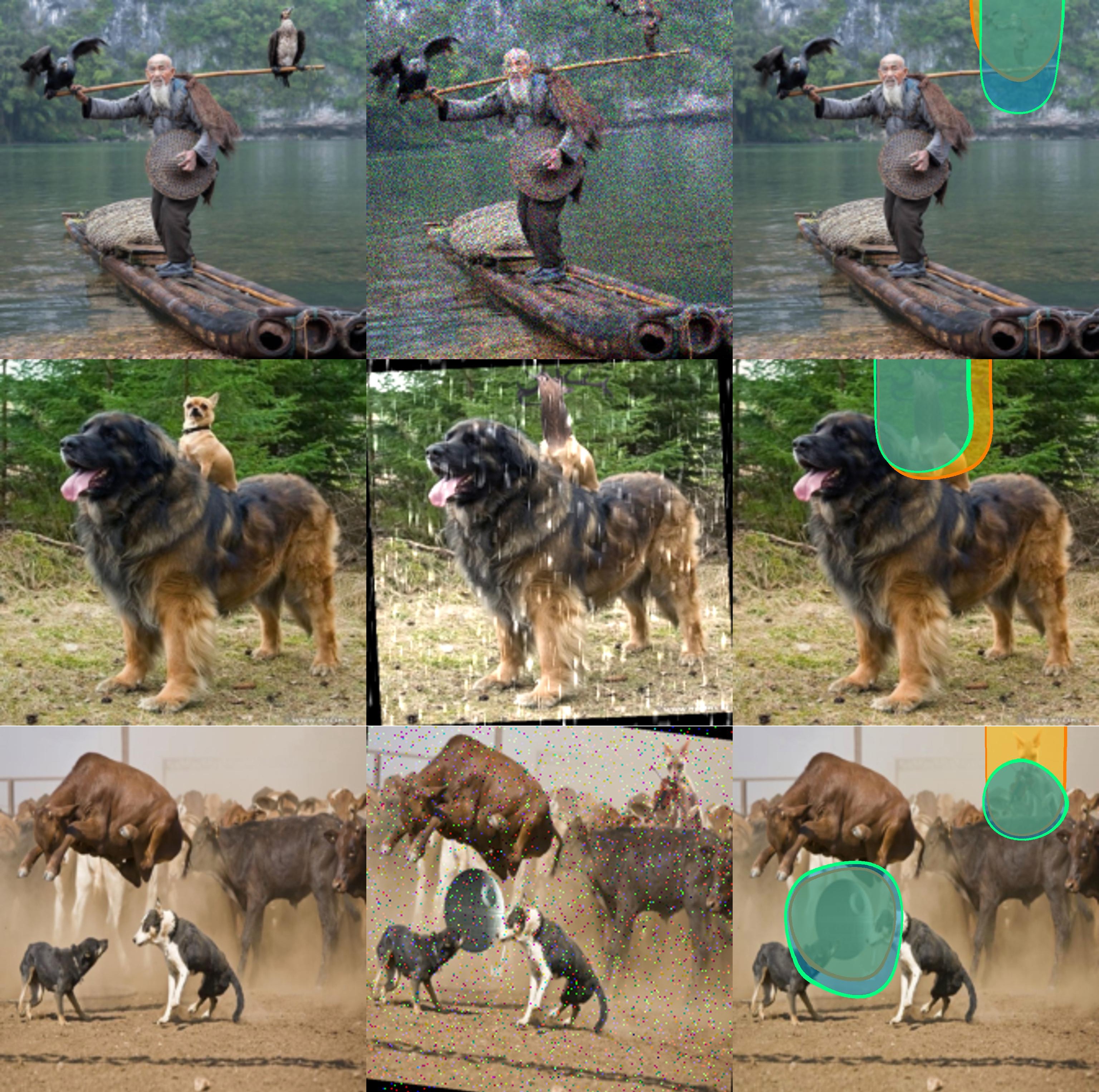}
    \caption{Heatmap results. Left col.: Original image.  Middle col.: Manipulated image also subjected to benign transformation.  Right col.: Heatmap output (green) ignoring benign transformation and highlighting manipulation (ground truth in yellow).}
    \squeeze
    \label{fig:repres}
\end{figure}

\begin{table}
    \caption{Evaluating heatmap accuracy and intrepretability for baseline methods.  Our proposed SImProv method is compared against baselines both objectively for accuracy (IoU) and subjectively via users to determine which exhibits best intrepretability (\% method preference). $\mathcal{F}_A$ indicates geometric alignment module applied.}
    \centering
    \resizebox{\columnwidth}{!}{%


    \begin{tabular}{l|cc}
        Method & \texttt{Accuracy (IoU)} & \texttt{Interp. (\%)} \\ \hline
        Ours & \bf 0.565 & \bf 84.6 \\
        ResNetConv+Geo. Align. $\mathcal{F}_A$  & 0.243 & 5.50 \\
        SSD+Geo. Align. $\mathcal{F}_A$ & 0.231 & 3.30 \\
        MantraNet+Geo. Align. $\mathcal{F}_A$ & 0.061 & 2.75 \\
    \end{tabular}}
    
    \squeeze
    \label{tab:iou}
\end{table}

The proposed method is compared with four baselines in terms of localization performance. The first baseline, \textbf{Sum of Squared Distances (SSD)}, calculates SSD between two images at the pixel level, resizes it to $7\times 7$, and then resizes it back before thresholding to create continuity in the detected heatmap. The second baseline, \textbf{ResNetConv}, extracts $7\times 7\times2048$ features from a pre-trained ImageNet ResNet50 model for both query and original images. These are averaged across channels to produce a $7\times7$ heatmap. \textbf{ErrAnalysis} - inspired from the blind detection technique in \citep{ela}, we perform JPEG compression on the query image and compare with itself. \textbf{MantraNet} - is a supervised blind detection method \citep{mantranet2019cvpr} that detects anomalous regions. Additionally we evaluate baselines with images passed through our alignment module.

We compare the heatmaps generated by our SImProv with baseline methods. Heatmaps are produced by upsampling the $7\times7$ heatmap output of the SImProv to the size of the image using bicubic interpolation.  Heatmaps may be presented on false-colour scale (\eg jet) in this form, or thresholded to produce an outline of the predicted manipulated region. In our experiments, we threshold the normalized heatmaps at 0.35 determined empirically.  Tab. \ref{tab:iou} (first column) reports the IoU metric between the predicted heatmap and the ground truth, both with and without the thresholding.  Whilst most baselines are improved through use of our geometric alignment ($\mathcal{F}_A$) process, our SImProv significantly exceeds baseline performances by at least 0.30. Change localization examples for SImProv are shown in Fig.~\ref{fig:repres}.

The effectiveness of heatmap interpretability is compared to baseline methods using a crowd-sourced study on Amazon Mechanical Turk (MTurk). Participants are shown an original image and an image that has been altered, along with the ground truth for the altered image. The altered image is also accompanied by a grid of heatmaps generated by nine different methods. The 9 methods included our own, 4 baselines (SSD, MantraNet, ErrAnalysis, and ResNetConv), and 4 warp-corrected baselines that used $\mathcal{F}_A$ for geometric alignment. Each of the 200 tasks was annotated by 5 different participants.

Tab.~\ref{tab:iou} (final col.) presents the results, which favor our proposed method, even when the image pair are pre-aligned. 

\squeezesm
\subsection{Large Scale Retrieval}

\begin{figure}[t!]
    \centering
    \includegraphics[width=0.85\linewidth]{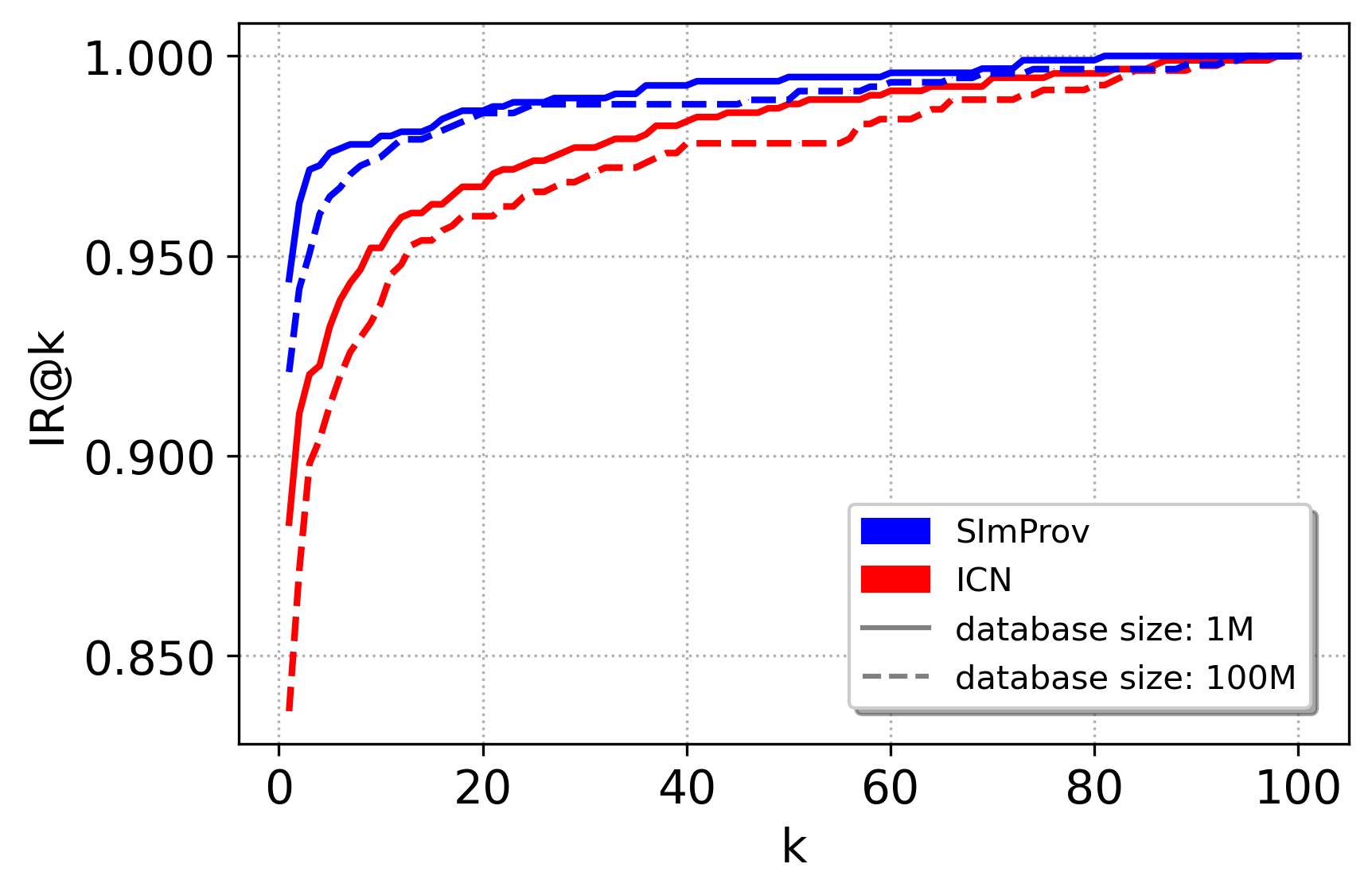}
    \caption{Retrieval performance comparison of ICN \citep{ICN} and SImProv on a 1M subset and fullset of BAM-100M, using BAM-Q-Aug queries.}
    \squeeze
    \label{fig:large_scale_1}
\end{figure}

\begin{figure}[ht]
    \centering
    \includegraphics[width=0.85\linewidth]{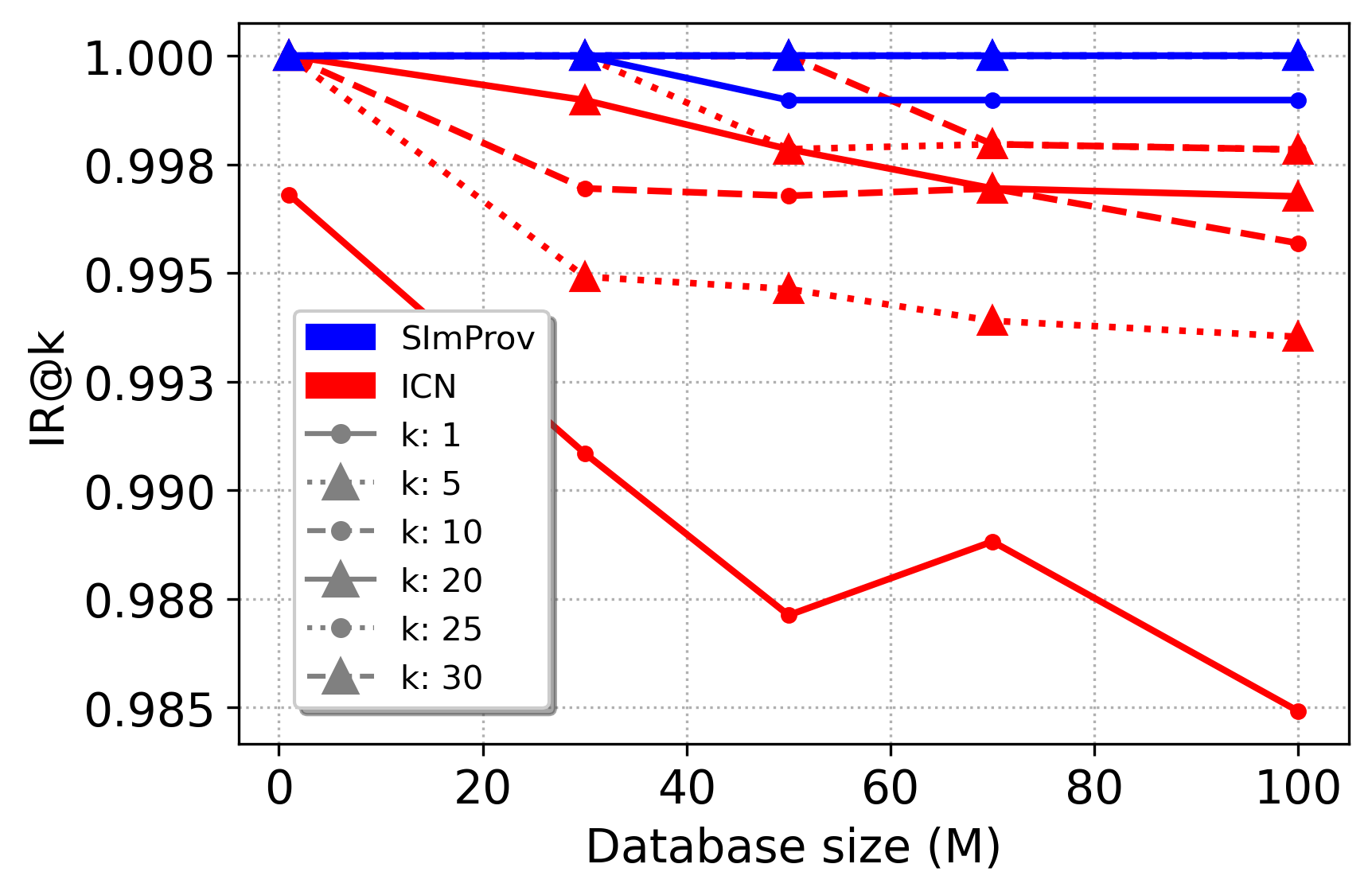}
    \caption{Top-k retrieval performance comparison of ICN \citep{ICN} and SImProv versus database size, using BAM-Q-Res queries.}
    \squeeze
    \label{fig:large_scale_2}
\end{figure}

We evaluate the scalability of our method by indexing the BAM-100M database. We compare the $IR@k$ performance of SImProv to its earlier version ICN \citep{ICN}. Fig. \ref{fig:large_scale_2} shows $IR@k$ versus database size curves of SImProv and ICN on BAM-100M with BAM-Q-Res as the query set. We demonstrates that SImProv's performance does not degrade nearly as much as ICN with increase in database size. For SImProv, the $IR@k$ remains nearly $1.0$ at all image database sizes, dipping to $0.999$ for the most challenging case of $IR@1$ for database sizes above 30M. ICN, on the other hand, is much greater affected by database size, with $IR@1$ dropping from $0.997$ at 1M images to $0.985$ at 100M images. Results for the more challenging query set BAM-Q-Aug are depicted in Fig.~\ref{fig:large_scale_1}. SImProv outperforms ICN by a large margin at early k values, on both BAM-100M and a subset of 1M images. The performance drop when increasing the database size from 1M to 100M for SImProv is also lower than ICN. The IR@k curves converge as k value reaches 100, and saturated performance is achieved at IR@100 for both methods regardless of database size, which justifies our design choice of selecting top-100 images for SImProv subsequent stages. 

\subsection{Evaluating Classification}

\begin{table}[t!]
\caption{SImProv stage 2 - Same / different classifier}
\centering
\begin{tabular}{lcc}
\multirow{2}{*}{\ttfamily Test} & \multicolumn{2}{c}{\ttfamily Average Precision (AP)} \\
& \ttfamily  ICN \citep{ICN} & \ttfamily  Ours \\
\hline \hline
Original      & 1.000  & 1.000 \\ 
Benign        & 0.9996 & 1.000 \\
Manip.        & 0.9976 & 0.9922 \\
Benign+Manip. & 0.9962 & 0.9909 \\
Distinct      & 0.9895 & 0.9973 \\
\hline
\end{tabular}
\label{tab:classification1}
\squeeze
\end{table}

\begin{table}[t!]
\caption{SImProv stage 3 - Benign / Manip classifier}
\centering
\begin{tabular}{lcc}
\multirow{2}{*}{\ttfamily Test} & \multicolumn{2}{c}{\ttfamily Average Precision (AP)} \\
& \ttfamily  ICN \citep{ICN} & \ttfamily  Ours \\
\hline \hline
Original      & 1.000  & 1.000 \\ 
Benign        & 0.9635 & 0.9800 \\
Manip.        & 0.9726 & 0.9932 \\
Benign+Manip. & 0.8807 & 0.9793 \\
\hline
\end{tabular}
\label{tab:classification2}
\squeeze
\end{table}

We evaluate the classification performance of two classifiers: same/different in SImProv stage 2 and benign/manipulated in stage 3. Same/different classification is an output of PSEN \ref{method:reordering}, which classifies a pair of images as either being two entirely different images, or the same image, potentially under different transformations. Benign/manipulated classification is an output of the change localization network \ref{method:loc}, which assumes that the input images are not distinct and focuses on classifying whether the differences between them are benign or editorial. We compare the performance of our approach with ICN \citep{ICN}, which has a single 3-way (benign, manipulated, distinct) classifier. In case of same/different evaluation, we combine the confidences of `benign' and `manipulated' to count as `same'.

We evaluate the performance of the 2-way same/different classification by comparing each original image in the test set with: itself, benign transformed version of itself, manipulated version, manipulated as well as benign transformed and an entirely different image, chosen at random. All of the cases except the last are expected to be classified as `same' and the last one as `distinct'.  Tab. \ref{tab:classification1} shows the Average Precision (AP) scores achieved for each case. A non-modified original-original pair is always correctly classified as the same image by both both methods. Introduction of benign transformations reduces the accuracy of ICN slightly, but does not affect our approach. The most challenging case is queries that are both manipulated and benign transformed, however both methods maintain AP near $0.99$ in all of the cases. 

The bigger difference in performance can be seen in Tab. \ref{tab:classification2} which shows the AP scores for benign/manipulated classification. Here, our approach outperforms ICN by $~2\%$ in the cases where the query image is either just benignly transformed or just manipulated. The difference in performance grows to $~9.9\%$ when the query is both manipulated and benign transformed.

\squeeze
\squeezesm
\section{Conclusion}

We presented a Scalable Image Provenance (SImProv) framework for large-scale retrieval and visual comparison of a pair of images in order to detect and localize manipulated regions. SImProv enables users to match images circulating `in the wild' to a trusted database of original images. When a query image is matched to an original, SImProv generates a heatmap that highlights areas of manipulation while ignoring benign transformations that can occur when images are shared online. We introduced two main architecture changes compared to an earlier version of the work \citep{ICN}: incorporation of instance-level feature pooling for image retrieval and combination of individual and pairwise descriptors for near-duplicate detection, followed by re-ranking. We show that feature pooling improves retrieval performance by enabling the use of queries of larger resolution. We use a large corpus of 100 million diverse images to demonstrate that these changes improve retrieval performance and make our approach applicable to the web-scale content authenticity problem.

\squeeze
\bibliographystyle{model2-names}
\bibliography{refs}

\end{document}